\pdfoutput=1

\documentclass[11pt]{article}

\usepackage[preprint]{acl}

\usepackage{times}
\usepackage{latexsym}

\usepackage[T1]{fontenc}

\usepackage[utf8]{inputenc}

\usepackage{microtype}

\usepackage{inconsolata}

\usepackage{graphicx}
\usepackage{algorithm}
\usepackage{algorithmic}
\usepackage{multirow}
\usepackage{booktabs}
\usepackage{amsmath}
\usepackage{subfigure}
\usepackage{ulem}
\usepackage{enumitem}
\usepackage{stfloats}
\usepackage{arydshln}
\title{ProgramTab: Boosting Table Reasoning of LLMs \\ via Programmatic Paradigm}

\author{Pei Guo, 
Enjie Liu, Yunzhi Tan\thanks{\; Corresponding Author}, Mochi Gao, Jianxin Zhang, \\
\textbf{Ruichao Zhong, Juntao Li, Bo Hu\footnotemark[1], Zang Li} \\
\textsuperscript{$\spadesuit$}Big Data and AI Platform Department, Tencent, China \\
\textsuperscript{$\diamondsuit$}Institute of Computer Science and Technology, Soochow University, China \\
\texttt{\{anthonyguo, karolinaliu, boristan, 
mochigao, harryyfhu, gavinzli\}@tencent.com}; \\
\texttt{20204027008@stu.suda.edu.cn}; \texttt{rzhongab@connect.ust.hk}; 
\texttt{ljt@suda.edu.cn} \\
 }

\begin{document}
\maketitle
\maketitle
\begin{abstract}
Table-based reasoning with large language models (LLMs), which requires reasoning based on natural language questions and structured tabular data, 
has gained widespread attention.
However, a series of issues still constrain the application of this task.
The previous approaches suffered from significant performance degradation when faced with large tables due to the difficulty of long text modeling and the limitation of input length for LLMs.
The text-to-SQL approach is used to efficiently extract key information from tables and generate smaller sub-tables. However, tabular data, especially web tables, often lack the necessary structure and consistency, making them unsuitable for performing mathematical logic operations using SQL queries.
We propose the ProgramTab framework, which guides LLMs employing in-context learning to perform tabular data preprocessing with Python code, as well as the momentous contents extraction with row and column extraction and SQL generation.
The experiment results on table reasoning datasets demonstrate that the ProgramTab framework effectively deals with table-based reasoning tasks and outperforms all LLM-based baselines.

\end{abstract}

\section{Introduction}
\label{sec:introduction}

Tables, as a popular form of data representation, play a significant role in everyday work and life. 
Analysis and reasoning based on tabular data have emerged as a hot topic in natural language processing, attracting wide attention from academia and industry. 
The main downstream tasks of tabular reasoning include table-based fact verification~\cite{chen2020tabfact, Aly2020feverous} and table-based question answering~\cite{pasupat-liang-2015-compositional,cho2015Explanatory}. 
The challenges of these tasks lie in how to enable language models to comprehend table data content, including text, numbers, etc., establish their connection with user queries, and execute efficient logical reasoning and computations.

\begin{figure}[t]
\centering
\includegraphics[scale=0.58]{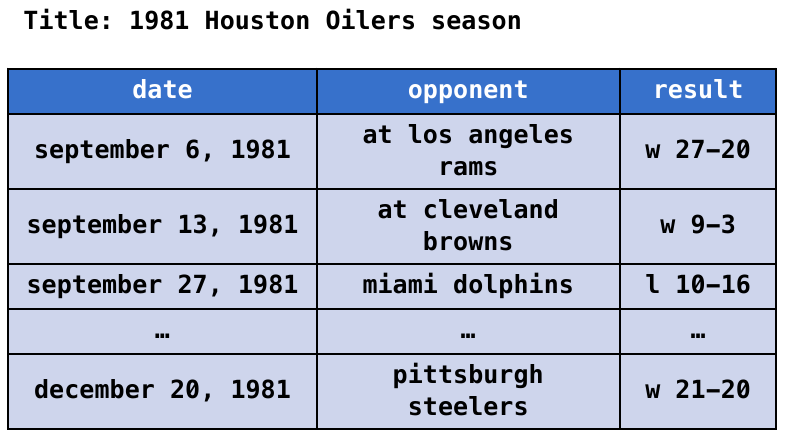}
\caption{An example of a table in WikiTQ dataset.}
\label{fig:table_sample}
\end{figure}

Recently, LLMs~\cite{brown2020,Hoffmann2022,gpt3.5} have significantly transformed the landscape of natural language processing tasks with their impressive understanding and generation capabilities. Instead of fine-tuning the pre-trained models, sufficiently utilizing the in-context learning of LLMs to solve complex tabular data reasoning has been a mainstream direction~\cite{chen-2023-large,Binder,ye2023large,wang2024chain}. 
However, current methods still face several limitations. Firstly, most of the work~\cite{Binder,ye2023large,wang2024chain} treats the entire table as an input, which is unsuitable for tables containing large amounts of data. When the number of tokens in a table exceeds the maximum input limitation of LLMs, the content of the table will be truncated, leading to information loss and consequently affecting the performance of LLMs. This has been verified in the work of~\cite{chen-2023-large}.
To mitigate the length constraint of inputs, the common approach is to utilize a programmatic language, such as generating SQL queries to retrieve the most relevant rows and column data~\cite{ye2023large,nahid2024tabsqlify,zhang-etal-2024-e5,alter}. 
However, \textbf{table data, especially the web table is usually provided as strings and often lacks the necessary structure and consistency, requiring conversion to the appropriate format and data types for mathematical logic operations to avoid calculation errors. }
It requires SQL to preprocess the data while extracting the relevant information, which increases the complexity of generating SQL (The results in Table~\ref{tb.study} can confirm this opinion).
For example, as shown in Figure~\ref{fig:table_sample}, 
when the question is about the number of games the Houston Oilers won in 1981 season, the ‘w’ and ‘l’ symbols from the result cell aren't provided as a single column and need to be extracted, which is defined as "lack of necessary structure". Regarding the absence of consistency, the structure at the "year" column in Figure~\ref{fig:framework} is inconsistent, such as "1931" and "spring 1932".

To address the above challenges, we introduce the ProgramTab framework, which executes with program languages (Python and SQL) to flexibly handle the table contents based on the questions.
Specifically, as shown in Figure~\ref{fig:framework}, (1) we utilize the embedding model to compute the relevant scores of each line of tables with the questions and resort the lines in descending order. 
In the following steps, the top K lines with higher relevant scores are extracted as instances to replace the complete tables.
With the most relevant lines as input, (2) LLMs are prompted to select the columns related to the questions, (3) generate the Python code to preprocess the table data, including unifying the data format and defining the data type for each column. 
After that, (4) SQL queries are generated using chain-of-thought (CoT)~\cite{wei-2023-cot} and executed to obtain the most valuable information. Finally, (5) LLMs process this information and produce the final answers.

We validate our ProgramTab framework by conducting experiments on two challenging table reasoning datasets: WikiTQ~\cite{pasupat-liang-2015-compositional} and TabFact~\cite{chen2020tabfact}. 
With three LLM backbones, our evaluation demonstrated that ProgramTab achieves excellent performance on table-based reasoning benchmarks, and outperforms all the other baselines with different LLM backbones. 
Besides, ProgramTab is not limited by the input length of table data, which obtains a significant efficiency and effectiveness improvement compared with other strong baselines.

\section{Related Work}
In this section, we introduce the related approaches of table-based reasoning: fine-tuning-based and prompting-based table reasoning.

\subsection{Fine-tuning-based Table reasoning}
Table-based understanding and reasoning tasks are significant in data analysis systems.
Many approaches focus on constructing pre-trained language models and fine-tuning them to address these tasks~\cite{zhang-etal-2020-table,patnaik2024cabinet}.
Among them, mask language models (MLM) are widely adopted. For example, 
TaPas~\cite{herzig-etal-2020-tapas} acquires BERT~\cite{devlin-etal-2019-bert} to parse table information via pre-training. 
PASTA~\cite{gu-etal-2022-pasta} pre-trains DeBERTaV3~\cite{he2023debertav} to perform six types of common sentence–table cloze tasks.
Besides, TAPEX~\cite{TAPEX} employs the BART~\cite{lewis-etal-2020-bart} model to learn the neural SQL executors over a synthetic corpus.
OmniTab~\cite{jiang-etal-2022-omnitab} leverages retrieval to pair relevant natural sentences with mask-based pre-training and synthesizes natural language questions by converting sampled SQL from tables.
Inner Table Retrieval (ITR)~\cite{lin-etal-2023-inner} extracts sub-tables to preserve the most relevant information for the questions.

\label{sec:programtab}
\begin{figure*}[t]
\centering
\includegraphics[scale=0.26]{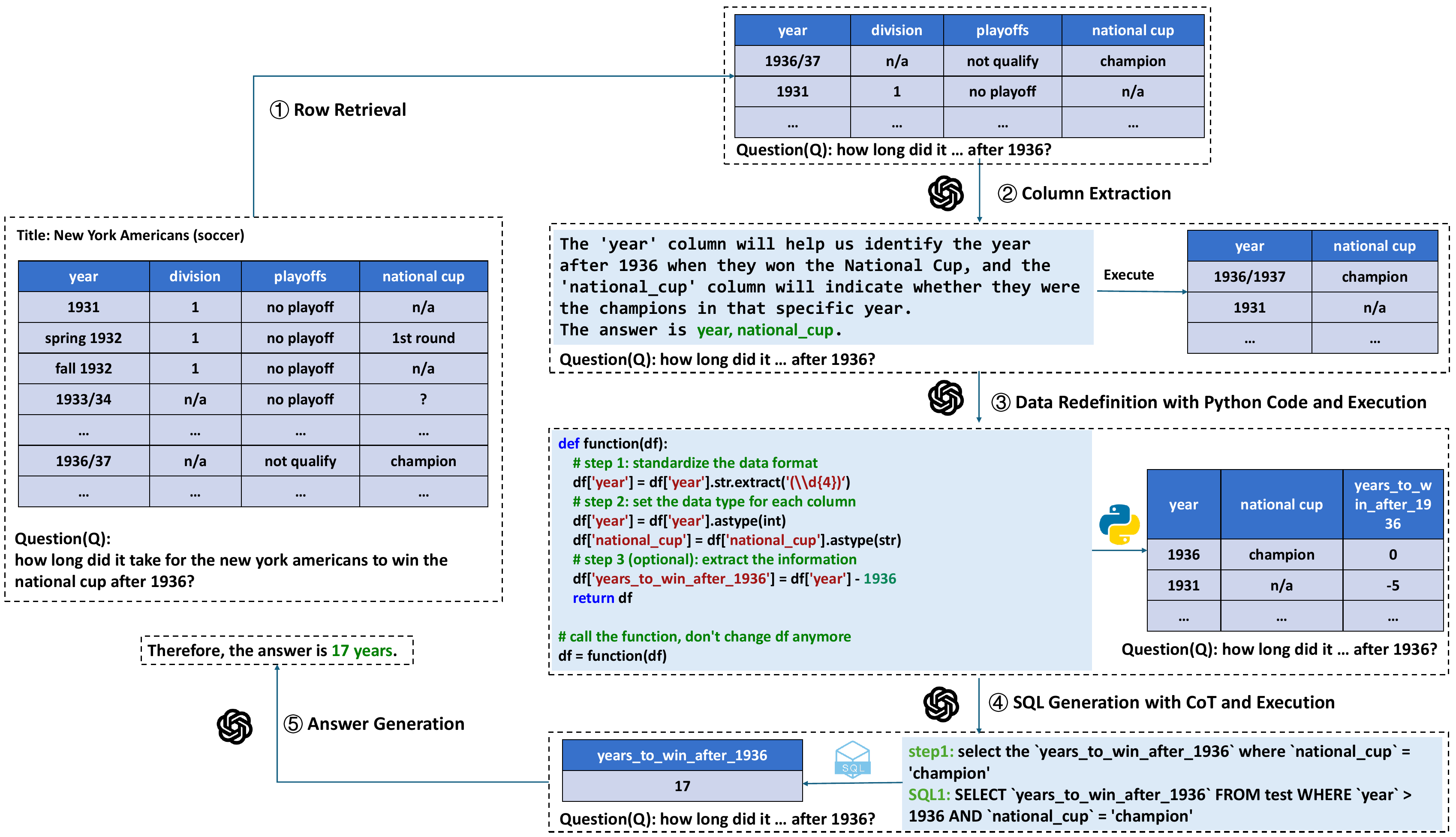}
\caption{The overview of ProgramTab for table-based reasoning.}
\label{fig:framework}
\end{figure*}

\subsection{Prompting-based Table Reasoning}
Recently, LLMs~\cite{Hoffmann2022,gpt4,llama} have gained widespread attention due to their powerful understanding and generation capabilities.
Given a few augmenting few-shot examples relevant to the tasks, the LLMs can tackle various reasoning tasks~\cite{Fu_cot,zhang_cot}.
A few approaches also employ LLMs to tackle table reasoning tasks with few-shot prompts.
TableCoT~\cite{tai-etal-2023-exploring} systematically explores the performance of LLMs 
on table reasoning tasks and finds that LLMs are excellent at solving such tasks, especially combined with CoT approach.
Besides, rather than generating general text, additional programmatic text, such as Python programs~\cite{pot,pal-gao-2023}, and Text-to-SQL~\cite{text2sql} approaches are employed to improve the performance further.
LEVER~\cite{Lever-2023} improves the performance of code LLMs on language-to-code tasks by training separate verifiers to validate the programs generated by LLMs and their execution results.
Binder~\cite{Binder} maps the task input to a program that allows generating SQL or Python programs and extending their functions by calling LLMs in the program.
ReAcTable~\cite{zhang-2024-reactable} breaks down the problem into multiple steps and uses LLMs to generate code programs that are executed through external tools for each step. Finally, it leverages majority voting to improve overall accuracy.
\citet{wang2024chain} proposes a Chain-of-Table framework that designs a series of table operations and dynamically plans an operation chain based on the inputs.
It's difficult for LLMs to perform reasoning when confronted with large tables with multiple rows.
Dater~\cite{ye2023large}, TabSQLify~\cite{nahid2024tabsqlify} and H-STAR~\cite{h-star} decompose the original table into the sub-table by selecting the relevant rows and columns. After that, Dater and Alter~\cite{alter} also propose the parsing-execution-filling and query augmentation strategy respectively to decompose a complex question into simpler step-by-step sub-questions by generating an intermediate SQL.
$E^5$~\cite{zhang-etal-2024-e5} presents an algorithm to condense large tables while maintaining useful information.
\textbf{Unlike above works, which typically extract information directly using SQL queries to obtain answers—thus increasing the difficulty of SQL generation, we propose an innovative approach that leverages LLMs to generate code for data preprocessing.} 
Besides, the most similar work is NormTab~\cite{nahid-rafiei-2024-normtab}, which utilizes LLMs to regularize table content, making it conducive to SQL query generation. 
\textbf{ProgramTab leverages LLMs to generate relevant processing code for table normalization, whereas NormTab has LLMs directly output the entire table in text form.}
In comparison, ProgramTab is more efficient, and through code execution, it reduces the likelihood of inconsistencies in the processing of data within the same column. 
\textbf{Additionally, we propose an improvement for SQL generation by decomposing the question into multiple sub-questions and providing the corresponding SQL for each sub-question, aiming to further enhance performance.}


\section{ProgramTab Reasoning}
\label{sec:programtab}
As shown in Figure~\ref{fig:framework}, ProgramTab consists of five procedures:
1) \textbf{row retrieval}, 2) \textbf{column extraction}, 3) \textbf{data definition with code}, 4) \textbf{SQL generation} and 5) \textbf{answer generation}.
In this section, we describe the above procedures in detail.
The original table is denoted as $T$.
\subsection{Row Retrieval}
\label{sec:row_retrieval}
To alleviate the limitation of the input length of LLMs, we first execute row retrieval, extracting the most relevant rows to represent the entire table content. 
Specifically, for each row of data in the table, we concatenate the column name and value of the cells to form a text segment, and an embedding model is utilized to calculate the relevance score between the row data and the question. 
Ultimately, the top K most relevant rows are selected as instances in the prompt templates of the following steps, which effectively alleviates the whole table as the input context.
The value of ‘K’ is determined based on the context length supported by the model being used. We set K to 10 for all the models. 

\begin{figure}[t]
\centering
\includegraphics[scale=0.42]{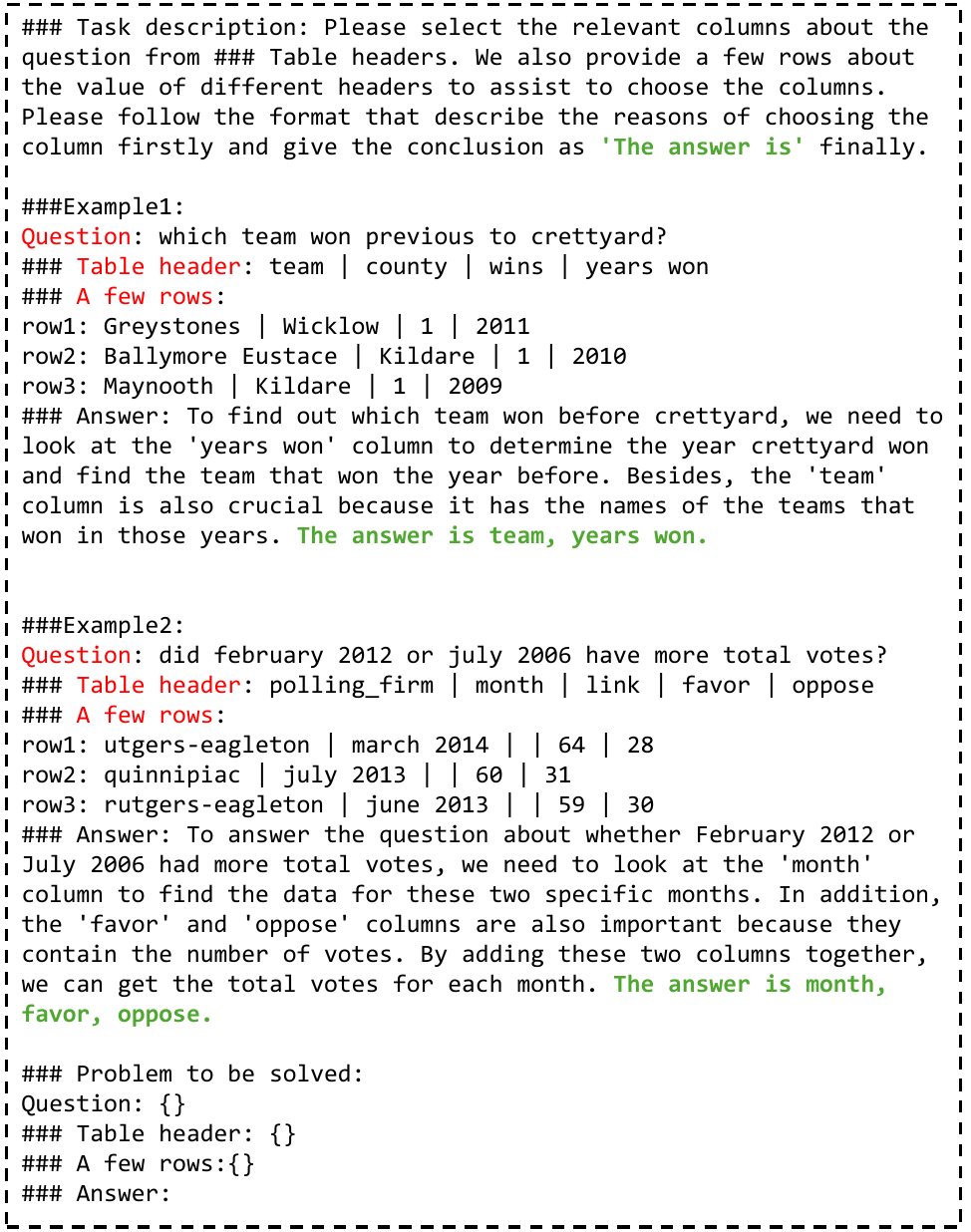}
\caption{Prompt for LLMs to extract columns.}
\label{fig:column_select}
\end{figure}

\subsection{Column Extraction}
\label{sec:column_extraction}
To minimize the impact of irrelevant data, it is essential to extract the relevant columns and utilize them for LLMs to conduct reasoning~\cite{alter}.
As shown in Figure~\ref{fig:column_select}, given the question, table header, and top K rows as input context, we prompt LLMs to follow the examples and extract the related columns with additional explanation.
Based on the LLMs filter columns, we extract them from $T$ and obtain $T_{col}$.

\subsection{Data Redefinition with Code}
\begin{figure}[t]
\centering
\includegraphics[scale=0.42]{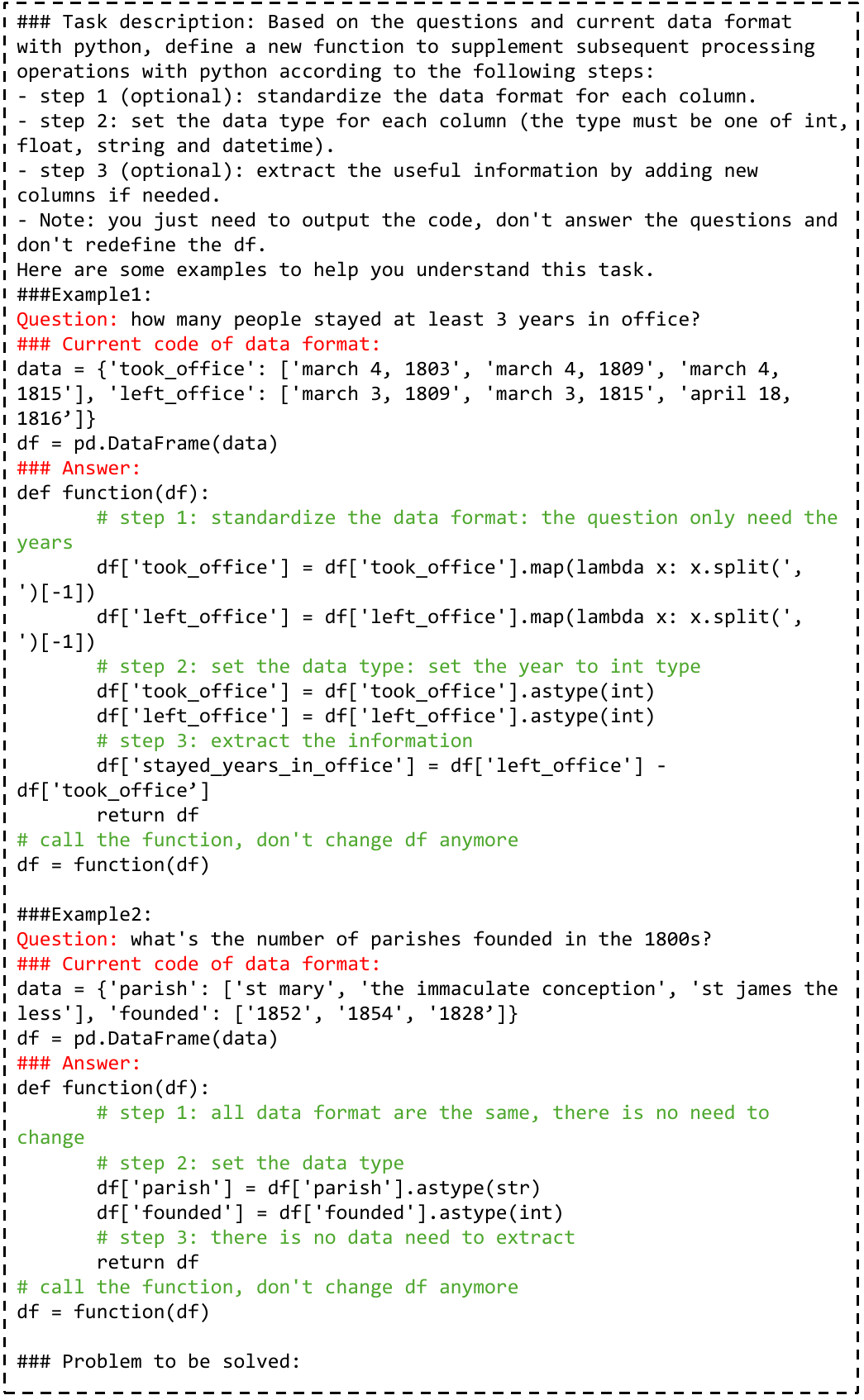}
\caption{The prompt for LLMs to perform data redefinition with code.}
\label{fig:data_define}
\end{figure}
\label{sec:data_redefinition}
Specifically, to maintain the flexibility of table data, the string type is adopted for the table data especially collected from the web. 
Besides, the format of data is not always consistent which causes a great challenge for SQL generation.
For example, as shown in Figure~\ref{fig:framework}, the values of column $year$ in $T$ are not rigorous, which conclude three different formats with string type: `1931', `spring 1932', and `1933/34'. 
Therefore, it's necessary to redefine data, including unifying the format, defining the data type, and extracting additional information (the detailed discussion about data redefinition is presented in Section~\ref{sec:study_operations}).
The related prompt is shown in Figure~\ref{fig:data_define}, given the current data format code, we acquire LLMs to generate Python code with the following steps. 
Firstly, if there is column data with inconsistent formats, standardize it to form a unified format.
Besides, the data type for each column must be set to make it suitable for performing mathematical logical operations, such as defining the column to the integer type.
We require that the data type must be one of integer, float, string, and datetime types.
Finally, additional information could be extracted by adding new columns.
Among them, annotations are added for each step to benefit LLMs to follow the above steps more effectively. 
Besides, steps 1 and 3 are optional, depending on the specific cases. 
For instance, the formats of each column in Figure~\ref{fig:data_define} example 2 are consistent, there is no need to extract extra information. As a result, steps 1 and 3 are unnecessary.


\subsection{SQL Generation}
\label{sec:sql_generation}

\begin{figure}[t]
\centering
\includegraphics[scale=0.42]{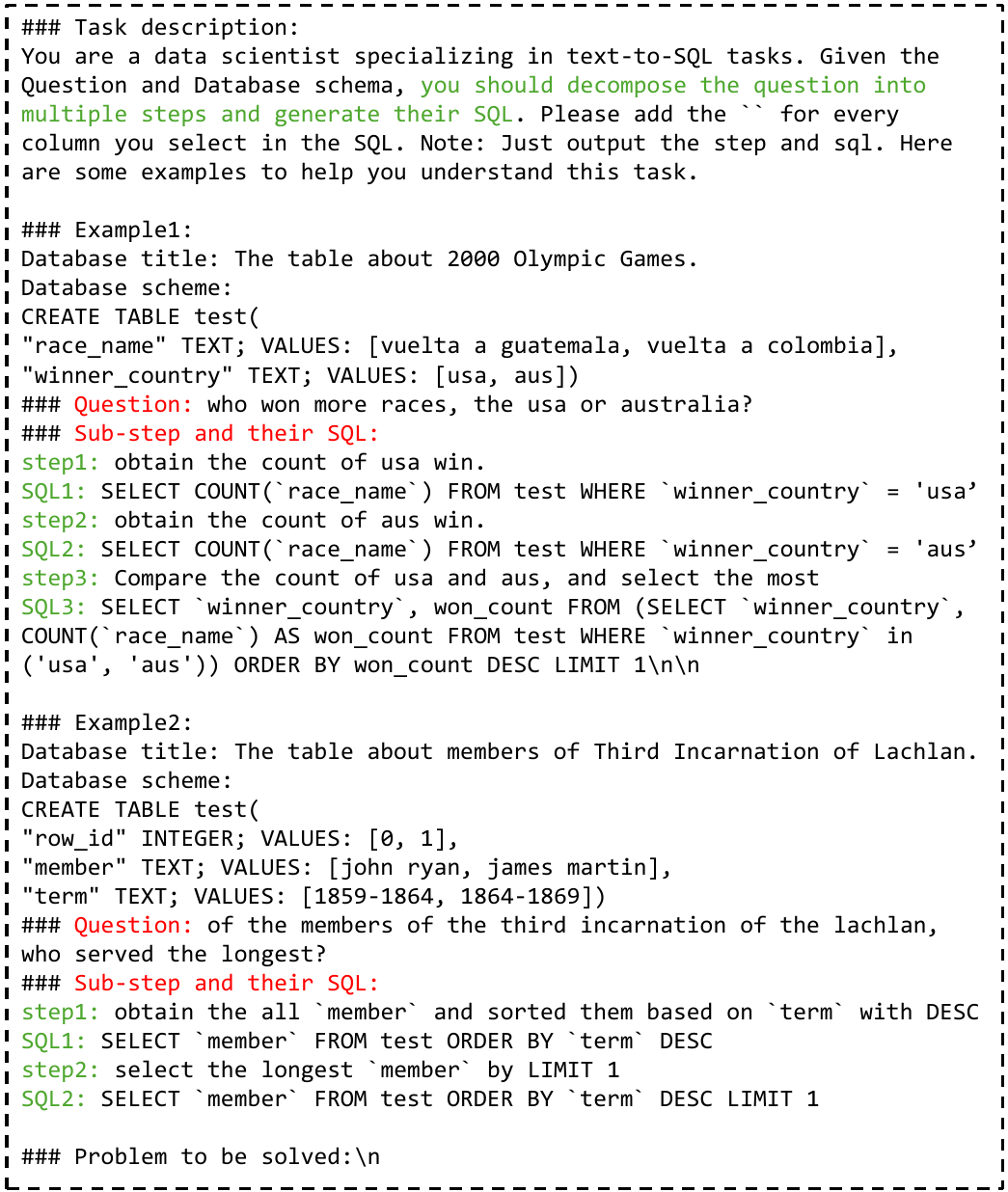}
\caption{Prompt for LLMs to generate SQL with CoT.}
\label{fig:sql_generation}
\end{figure}

The table data after redefining is unified and meets the requirements for SQL execution.
In this step, we make use of few-shot learning to prompt LLMs to perform SQL generation.
Specifically, as presented in Figure~\ref{fig:sql_generation}, the essential information is provided, such as the database title, schema, and top K relevant rows. 
With these contexts, LLMs are prompted to decompose the question into multiple steps and generate their sub-SQL with the CoT method.
We find that the CoT style is beneficial for LLMs to generate the final SQL queries, and the specific analysis is described in Section~\ref{sec:study_operations}.

\subsection{Answer Generation}

\begin{figure}[t]
\centering
\includegraphics[scale=0.42]{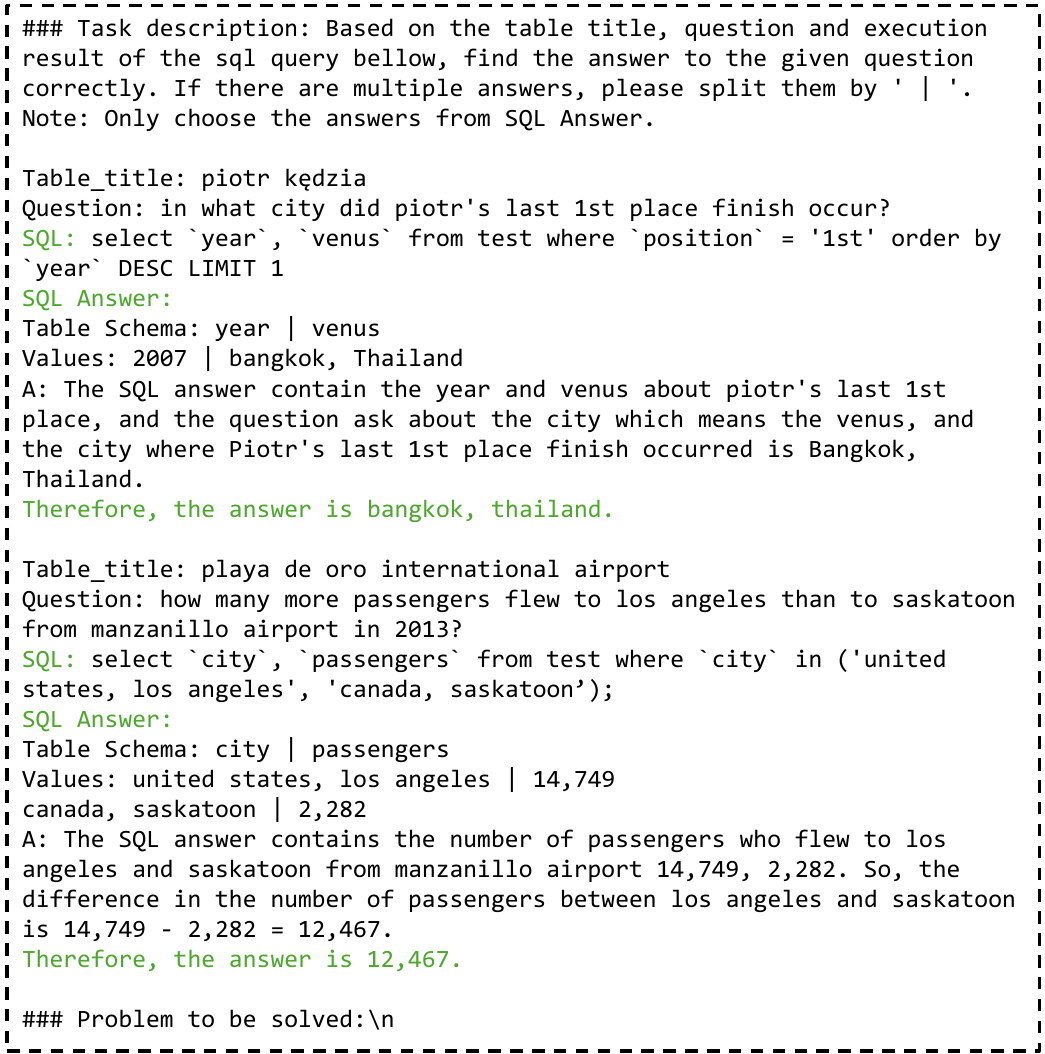}
\caption{Prompt for LLMs to generate final answers.}
\label{fig:answer_generation}
\end{figure}

After executing the SQL query obtained from the previous step, the most relevant information is gained from the table.
As presented in Figure~\ref{fig:answer_generation}, 
during this step, based on the results from executing the SQL query and the question, we utilize LLMs to reason with the additional explanation and finally make a conclusion.
Consequently, we can conveniently extract the results from the conclusions as final answers.
This approach helps LLMs concentrate on the relevant parts to understand the context and answer the questions.

\section{Experiments}

\subsection{Datasets}
We design relevant prompts and utilize the powerful in-context learning ability of LLMs to directly reason on the test set.
We evaluate the proposed ProgramTab on three public table reasoning benchmarks: TabFact~\cite{chen2020tabfact}, WikiTQ~\cite{pasupat-liang-2015-compositional} and HiTab~\cite{cheng-etal-2022-hitab}. 
Among them, TabFact is a table-based binary fact verification benchmark. Given a statement, we need to ascertain the truthfulness of it based on the table. We report the accuracy of the test set, which contains 2,024 statements and 298 tables.
Besides, WikiTQ is one of the most commonly
used and highly complex datasets, collected and annotated based on Wikipedia tables. The WikiTQ comprises 4,344 question-answer pairs in the test set.
HiTab is the dataset that contains hierarchical tables with complex hierarchical indexing.

\subsection{Baselines}
We divide the baselines into two categories: (1) approaches that spend additional computing resources
to train proprietary models with custom training data, such as 
TaPas~\cite{herzig-etal-2020-tapas}, GraPPa~\cite{yu2021grappa}, TAPEX~\cite{TAPEX}, PASTA~\cite{gu-etal-2022-pasta}, TaCube~\cite{zhou-etal-2022-tacube}, OmniTab~\cite{jiang-etal-2022-omnitab}, ITR~\cite{lin-etal-2023-inner} and CABINET~\cite{patnaik2024cabinet}.
(2) without training, approaches that design few shot prompts and employ the in-context ability of LLMs, such as TableCoT~\cite{tai-etal-2023-exploring}, ReAcTable~\cite{zhang-2024-reactable}, Binder~\cite{Binder}, Dater~\cite{ye2023large}, Chain-of-Table~\cite{wang2024chain}, Alter~\cite{alter}, $E^5$~\cite{zhang-etal-2024-e5}, NormTab~\cite{nahid-rafiei-2024-normtab}, TabSQLify~\cite{nahid2024tabsqlify} and H-STAR~\cite{h-star}.

\subsection{Implementation Details}
In our settings, we conduct experiments by utilizing closed-source LLMs (GPT-3.5-Turbo and GPT-4o-mini\footnote{https://openai.com/index/gpt-4o-mini-advancing-cost-efficient-intelligence/}) and the open-source LLM Llama-3.1-70B-Instruct\footnote{https://ai.meta.com/blog/meta-llama-3-1/} as the backbones.
The prompt templates for each procedure are described in Section~\ref{sec:programtab}.
Besides, the details of hyper-parameters are presented in Appendix~\ref{sec:appendix_hyper}. Notably, syntax errors occasionally occurred during data redefinition and SQL generation, resulting in non-executable code. To address this issue, we adopted a retry mechanism. Specifically, when a runtime error occurred during both processes, we attempted to rerun the process, with a maximum of five attempts. If all five attempts failed, it was concluded that LLMs were unable to handle the given table, and no further steps were executed.
About the evaluation metrics, we follow ~\citet{nahid2024tabsqlify} to use the official denotation accuracy and employ the binary classification accuracy for WikiTQ and TabFact datasets evaluation respectively.
GTE-base~\cite{li2023gte} is employed as the embedding model.

\begin{table}[t]
\centering
\small
\renewcommand\arraystretch{1.0}
\resizebox{0.48\textwidth}{!}{
\begin{tabular}{cccc}
\toprule
\textbf{Methods} & \textbf{Backbone} & \textbf{Accuracy} \\
\midrule
& \textit{Previous Work with Training} & \\
\hdashline[0.5ex/5pt]
TaPas & BERT & 83.9 \\
Tapex & BART & 86.7 \\
PASTA & DeBERTaV3 & \textbf{90.8} \\
\midrule
& \textit{Previous Work without Training} & \\
\hdashline[0.5ex/5pt]
E$^5$ & GPT-4 & 88.7 \\
\hdashline[0.5ex/5pt]
ReAcTable & \multirow{9}{*}{GPT-3.5-Turbo} & 73.1 \\
TableCoT &  & 73.1 \\
Binder &  & 79.1 \\
Dater &  & 78.0 \\
Alter & & \underline{84.3} \\
NormTab & & 68.9 \\
TabSQLify & & 79.5  \\
H-STAR & & 85.0 \\
\textbf{ProgramTab (Ours)} & & \textbf{85.9} \\
\hdashline[0.5ex/5pt]
Binder & \multirow{5}{*}{Llama-3.1-70B-Instruct} & 78.1 \\
Dater & & 81.6 \\
Chain-of-Table & & \underline{85.6} \\
TabSQLify & & 70.7 \\
\textbf{ProgramTab (Ours)} &  & \textbf{86.8} \\
\hdashline[0.5ex/5pt]
Binder & \multirow{5}{*}{GPT-4o-mini} & \underline{84.6} \\
Dater &  & 80.9 \\
Chain-of-Table & & 84.2 \\
TabSQLify & & 78.7 \\
H-STAR & & 89.4 \\
\textbf{ProgramTab (Ours)} & & \textbf{89.6} \\
\bottomrule
\end{tabular}
}
\caption{Accuracy of ProgramTab compared to the baselines on TabFact test set.}
\label{tb.main_results_tabfact}
\end{table}

\begin{table}[t]
\centering
\small
\renewcommand\arraystretch{1.0}
\resizebox{0.48\textwidth}{!}{
\begin{tabular}{cccc}
\toprule
\textbf{Methods} & \textbf{Backbone} & \textbf{Accuracy} \\
\midrule
& \textit{Previous Work with Training} & \\
\hdashline[0.5ex/5pt]
TaPas & BERT & 48.8 \\
GraPPa & RoBERTa & 52.7 \\
\hdashline[0.5ex/5pt]
Tapex & \multirow{5}{*}{BART} & 57.5 \\
TaCube &  & 60.8 \\
OmniTab & & 62.8 \\
ITR & & 63.4 \\
CABINET & & \underline{69.1} \\
\midrule
& \textit{Previous Work without Training} & \\
\hdashline[0.5ex/5pt]
TableCoT & \multirow{4}{*}{Codex} & 48.8 \\
Binder & & 61.9 \\
ReAcTable &  & 65.8 \\
Dater & & \underline{65.9} \\
\hdashline[0.5ex/5pt]
E$^5$ & GPT-4 & 65.5 \\
\hdashline[0.5ex/5pt]
ReAcTable & \multirow{9}{*}{GPT-3.5-Turbo} & 52.5 \\
TableCoT &  & 52.4 \\
Binder &  & 55.4 \\
Dater &  & 52.8 \\
Alter & & \underline{67.4} \\
TabSQLify &  & 64.7 \\
NormTab & & 61.2 \\
H-STAR & & 69.6 \\
\textbf{ProgramTab (Ours)} & & \textbf{70.3} \\
\hdashline[0.5ex/5pt]
Binder & \multirow{5}{*}{Llama3.1-70B-Instruct} & 50.5 \\
Dater & & 43.5 \\
Chain-of-Table & & \underline{62.2} \\
TabSQLify & & 55.8 \\
\textbf{ProgramTab (Ours)} &  & \textbf{75.5} \\
\hdashline[0.5ex/5pt]
Binder & \multirow{5}{*}{GPT-4o-mini} & \underline{58.8} \\
Dater &  & 58.3 \\
Chain-of-Table & & 55.6 \\
TabSQLify & & 57.0 \\
H-STAR & & 74.9 \\
\textbf{ProgramTab (Ours)} & & \textbf{76.0} \\
\bottomrule
\end{tabular}
}
\caption{Performance of ProgramTab compared to the baselines on WikiTQ test set.}
\label{tb.main_results_wtq}
\end{table}

\subsection{Results}
As presented in Table~\ref{tb.main_results_tabfact} and Table~\ref{tb.main_results_wtq} (the additional results on HiTab in Appendix~\ref{sec:appendix_hitab}.), 
(1) the previous work, training with specific tasks perform well. Specifically, PASTA~\cite{gu-etal-2022-pasta} achieves 90.8\% accuracy on TabFact, while CABINET~\cite{patnaik2024cabinet} obtains 69.1\% on WikiTQ. 
Using GPT-4o-mini as the backbone, ProgramTab achieved performance comparable to PASTA on the TabFact dataset. Furthermore, on the WikiTQ dataset, ProgramTab outperformed CABINET regardless of the large model used as its backbone.
Due to unnecessary additional fine-tuning,
the generalization of ProgramTab is better.
(2) Compared to previous work without training, ProgramTab with different LLM backbones outperforms the other baselines on all evaluation benchmarks. 
In addition, our framework with GPT-4o-mini achieves better performance compared to $E^5$ with GPT-4.
(3) With stronger coding and reasoning abilities, ProgramTab with Llama-3.1-70B-Instruct and GPT-4o-mini achieve better performance.
\section{Analysis}

\subsection{Ablation Study Results}
\label{sec:study_operations}
To evaluate the effectiveness of each procedure in the ProgramTab framework, we pay attention to two important steps: \textbf{d}ata \textbf{r}edefinition (DR) and \textbf{S}QL \textbf{g}eneration (SG).
Specifically, we remove the DR procedure described in Section~\ref{sec:data_redefinition} and keep the other steps unchanged.
The result in Table~\ref{tb.study} shows that without the DR step to preprocess the tabular data, it will require SQL to preprocess the data and extract the relevant information, which increases the complexity of generating SQL for LLMs.
Therefore, the performance significantly decreases on both datasets, especially on WikiTQ which is more complex.
This conclusion is also verified by \citet{wang2024chain}.
Besides, we replace the procedure described in Section~\ref{sec:sql_generation} with the SQL generation without CoT (denotes as w/o CoT SG).
The special prompt is shown in Appendix~\ref{sec:appendix_sql}.
Table~\ref{tb.study} presents that the performance of ProgramTab w/o SG CoT drops when discarding question decomposition.
It verifies that compared with direct SQL generation, decomposing the questions into multiple steps and generating their sub-SQL is effective in reducing the difficulty of SQL generation.

\begin{table}[t]
\centering
\small
\renewcommand\arraystretch{1.0}
\resizebox{0.48\textwidth}{!}{
\begin{tabular}{lcc}
\toprule
\textbf{Methods} & \textbf{TabFact} & \textbf{WikiTQ} \\
\midrule
Binder & 79.1 & 55.4 \\
Dater & 78.0 & 52.8 \\
TabSQLify & 79.5 & 64.7 \\
ProgramTab & \textbf{85.9} & \textbf{70.3} \\
\quad w/o DR & 81.6 ($\downarrow$ 4.3) & 59.4 ($\downarrow$ 10.9) \\
\quad w/o CoT SG & 84.1 ($\downarrow$ 1.8)  & 65.0 ($\downarrow$ 5.3) \\
\bottomrule
\end{tabular}
}
\caption{Ablation results of GPT-3.5-Turbo-based ProgramTab with and without data redefinition and CoT SQL generation.}
\label{tb.study}
\end{table}

\subsection{Performance Analysis under Large Tables}

As described in Section~\ref{sec:introduction}, ~\citet{chen-2023-large} and ~\citet{ye2023large} have presented that LLMs suffer from significant performance degeneration when dealing with large tables.
To evaluate the effectiveness of ProgramTab, we extract the large tables from WikiTQ and TabFact datasets. 
Specifically, we define the large tables for WikiTQ when the token counts are larger than 4000 because 4000 tokens are the maximum token limitation for GPT-3.5-Turbo. 
Besides, We follow ~\citet{nahid2024tabsqlify} to choose 1200 tokens for TabFact because the tables almost contain few data.
We then compare ProgramTab with Binder, Dater, Chain-of-Table, TableCoT, and TabSQLify.
As shown in Table~\ref{tb.large_table}, we observe that all strong baselines suffer from a significant decline in performance on two datasets.
For example, Binder with Codex merely achieves 29.6\% accuracy on the WikiTQ dataset and even can't be applied when utilizing GPT-3.5-Turbo as the backbone. 
Besides, TabSQLify obtains suboptimal performance thanks to its effective extraction of columns and rows employing the text-to-SQL method.
In contrast, ProgramTab significantly outperforms all baselines and even improves compared with performance on the full TabFact dataset.
It could be clarified that the row retrieval and column extraction procedures in our framework are effective in providing the relevant rows as the context, which is beneficial for SQL generation to extract the final information from the large tables.

\begin{table}[t]
\centering
\small
\renewcommand\arraystretch{1.2}
\resizebox{0.48\textwidth}{!}{
\begin{tabular}{lccc}
\toprule
\textbf{Methods} & \textbf{Backbone} & \textbf{TabFact} & \textbf{WikiTQ} \\
\midrule
Binder & Codex & - & 29.6 \\
Chain-of-Table & GPT-3.5-Turbo-16k-0613 & - & 44.8 \\
\midrule
Binder & \multirow{5}{*}{GPT-3.5-Turbo} & - & 0.0 \\
Dater & & - & 34.6 \\
TableCoT & & 55.5 & 35.1 \\
TabSQLify & & \underline{72.8} & \underline{52.3} \\
\textbf{ProgramTab} & & \textbf{86.6} & \textbf{68.0} \\
\bottomrule
\end{tabular}
}
\caption{Performance of ProgramTab and strong baselines on large tables from TabFact and WikiTQ.}
\label{tb.large_table}
\end{table}

\subsection{Robustness Analysis}

\begin{table}[t]
\centering
\small
\renewcommand\arraystretch{1.2}
\resizebox{0.48\textwidth}{!}{
\begin{tabular}{lccccc}
\toprule
\multirow{2}{*}{\textbf{Methods}} & \multirow{2}{*}{\textbf{Datasets}} & \multicolumn{4}{c}{\textbf{Cut-off(\%)}} \\
\cline{3-6}
& & \textbf{0-10\%} & \textbf{10-25\%} & \textbf{25-50\%} & \textbf{50\%+} \\
\midrule
TabSQLify & \multirow{2}{*}{WikiTQ} & 64.6 & \textbf{60.6} & \textbf{66.3} & 56.2 \\
ProgramTab & & \textbf{70.8} & \textbf{62.4} & 62.6 & \textbf{68.0} \\
\midrule
TabSQLify & \multirow{2}{*}{TabFact} & 79.1 & 80.8 & 70.0 & 72.8 \\
ProgramTab & & \textbf{89.0} & \textbf{86.5} & \textbf{77.5} & \textbf{86.4} \\
\bottomrule
\end{tabular}
}
\caption{Performance of ProgramTab on the different cutoff thresholds categories.}
\label{tb.cut_off}
\end{table}

Following \citet{nahid2024tabsqlify}, we verify the robustness of ProgramTab based on the different cutoff thresholds.
Specifically, the cutoff thresholds are established to discard tabular tokens exceeding these limits. 
For example, if the original table has 800 tokens and the maximum threshold is set to 600, it means that 200 tokens of the original table are truncated, and the percentage is 200/800 = 25.0\%.
In our experiment, we set the cutoff threshold at 2000 and 600 for WikiTQ and TabFact respectively. 
Table~\ref{tb.cut_off} shows four categories based on the above thresholds and presents that ProgramTab with GPT-3.5-Turbo outperforms TabSQLify except on the 25\%-50\% cutoff on WikiTQ.
The results further demonstrate that the ProgramTab can extract the relevant information under limited token boundary conditions and is not sensitive to input length limitations for LLMs.



\subsection{Efficiency Analysis}

\begin{table}[t]
\centering
\small
\renewcommand\arraystretch{1.2}
\resizebox{0.48\textwidth}{!}{
\begin{tabular}{lcc}
\toprule
\textbf{Methods} & \textbf{\# of samples / step} & \textbf{Total \# of samples} \\
\midrule
Binder & Neural SQL: 50 & 50 \\
\midrule
\multirow{4}{*}{Dater} & Decompose Table: 40 & \multirow{4}{*}{100} \\
& Generate Cloze: 20 & \\
& Generate SQL: 20 & \\
& Query: 20 & \\
\midrule
\multirow{3}{*}{Chain-of-Table} & Dynamic Plan $\leq$ 5 & \multirow{3}{*}{$\leq$ 25} \\
& Generate Args $\leq$ 19 &  \\
& Query: 1 & \\
\midrule
\multirow{2}{*}{TabSQLify} & Decompose Table: 1 & \multirow{2}{*}{2} \\
& Query: 1 & \\
\midrule
\multirow{4}{*}{\textbf{ProgramTab}} & Column Extraction: 1 & \multirow{4}{*}{4} \\
& Data Redefinition with Code: 1 & \\
& Generate SQL: 1 & \\
& Query: 1 & \\
\bottomrule
\end{tabular}
}
\caption{The number of samples generated by different methods adopting LLMs.}
\label{tb.samples_statistic}
\end{table}

Following \citet{wang2024chain}, 
we analyze the efficiency of ProgramTab from two aspects.
(1) We evaluate the number of samples generated by LLMs. 
For each reasoning step, compared to the approaches that apply the self-consistency (Binder and Dater) strategy to generate multiple samples or adopt the iterative sample creation process (Chain-of-Table), ProgramTab adopts a greedy search strategy to produce a single response. 
Specifically, Table~\ref{tb.samples_statistic} shows the number of samples generated by LLMs for a single question in different methods on the WikiTQ dataset.
We can find that LLMs are required to generate multiple samples for Binder and Dater, while Chain-of-Table adopts a more efficient approach to reduce the number of samples.
TabSQLify achieves the minimum number of samples.
Our approach adopts a greedy search strategy to obtain one response for each step, for a total of only four samples.
(2) We also validated the end-to-end performance of ProgramTab on the entire test set.
For each sample, ProgramTab requires an average time of 4 seconds to generate the final answer. In comparison, Dater requires generating 100 samples, with an average time of over 10 seconds. Clearly, our method has significantly lower latency.
Consequently, ProgramTab efficiently reduces computation time and resource costs and performs better.

\subsection{Error Analysis}

To systemically analyze the shortcomings of programTab with GPT-3.5-Turbo, we select two test sets (i.e., TabFact, and WikiTQ), and randomly choose 100 error samples from each dataset. 
Then, we manually examine these failures and they are classified into four error categories:
1) Missing Columns Error: LLMs don't select the relevant columns.
2) SQL Error: the generated SQL queries incorrectly filter the relevant information or contain syntax rule errors.
3) Code Error: the generated Python codes fail to unify the format and type of data, or introduce irrelevant information.
4) Reasoning Error: LLMs fail to generate the correct answers given the extracted relevant information.
It is worth noting that we additionally verify the performance of row retrieval in Appendix~\ref{sec:appendix_row}.
As shown in Figure~\ref{fig:error_analysis}, 
we can observe that the missing column and reasoning errors respectively account for a small portion of TabFact and WikiTQ.
The main source of errors focuses on the code and SQL errors, especially on the WikiTQ.
We replaced GPT-3.5-Turbo with GPT-4o-mini for code and SQL generation, and found that GPT-4o-mini effectively avoids the errors encountered with GPT-3.5-Tubo. The performance of these two LLMs in Table~\ref{tb.main_results_tabfact} and \ref{tb.main_results_wtq} can also be verified.
Consequently, enhancing the capacity of code generation is effective in improving the performance further. 

\begin{figure}[t]
\centering
\includegraphics[scale=0.35]{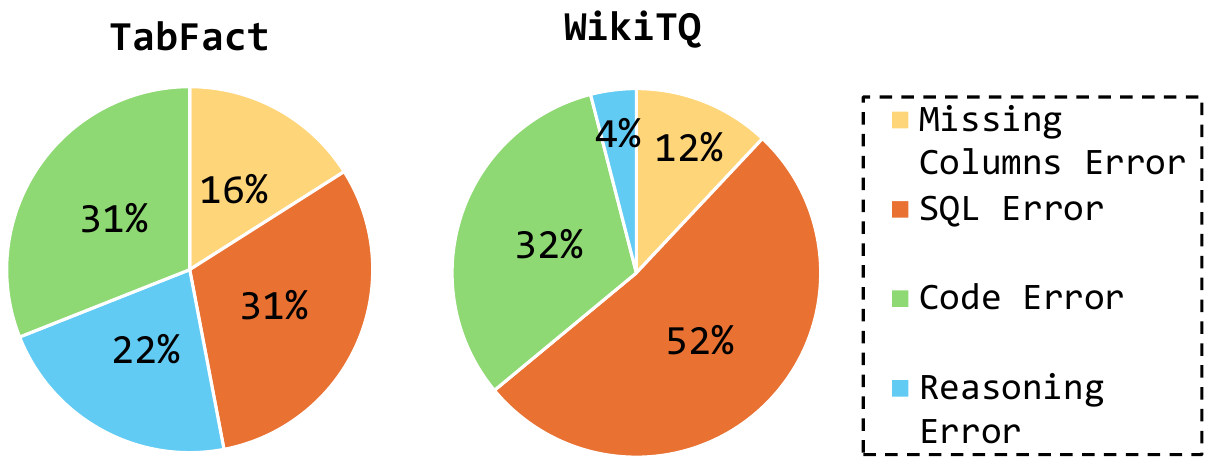}
\caption{Statistic of different error types on TabFact and WikiTQ datasets.}
\label{fig:error_analysis}
\end{figure}


\section{Conclusion}

In this paper, we illustrate the limitations of current table-based reasoning with LLMs approaches, including suffering from significant performance degradation when faced with large tables, and the inconsistent table data structure increases the difficulty of SQL generation.
Consequently, we propose the ProgramTab framework, which sufficiently implements the strong in-context learning ability of LLMs to perform tabular data preprocessing with Python code and key information extraction with SQL generation. 
It achieves the best performance compared 
with the baselines and is not limited by the input length of table data.
Hoping this flexible table-based reasoning framework can shed new light on the understanding of prompting LLMs for table understanding.
\section{Limitations}

In this section, we present several of the limitations of our approach - ProgramTab.
Firstly, the data redefinition with code can preprocess the table data well, but more preprocessing for more complex table structures should be explored further. 
What's more, how to perform row retrieval more efficiently from tables with large amounts of rows is another optimization direction.
\bibliography{custom}

\clearpage
\appendix

\section{Appendix}

\subsection{Prompts for SQL Generation without CoT}
\label{sec:appendix_sql}

\begin{figure}[h]
\centering
\includegraphics[scale=0.35]{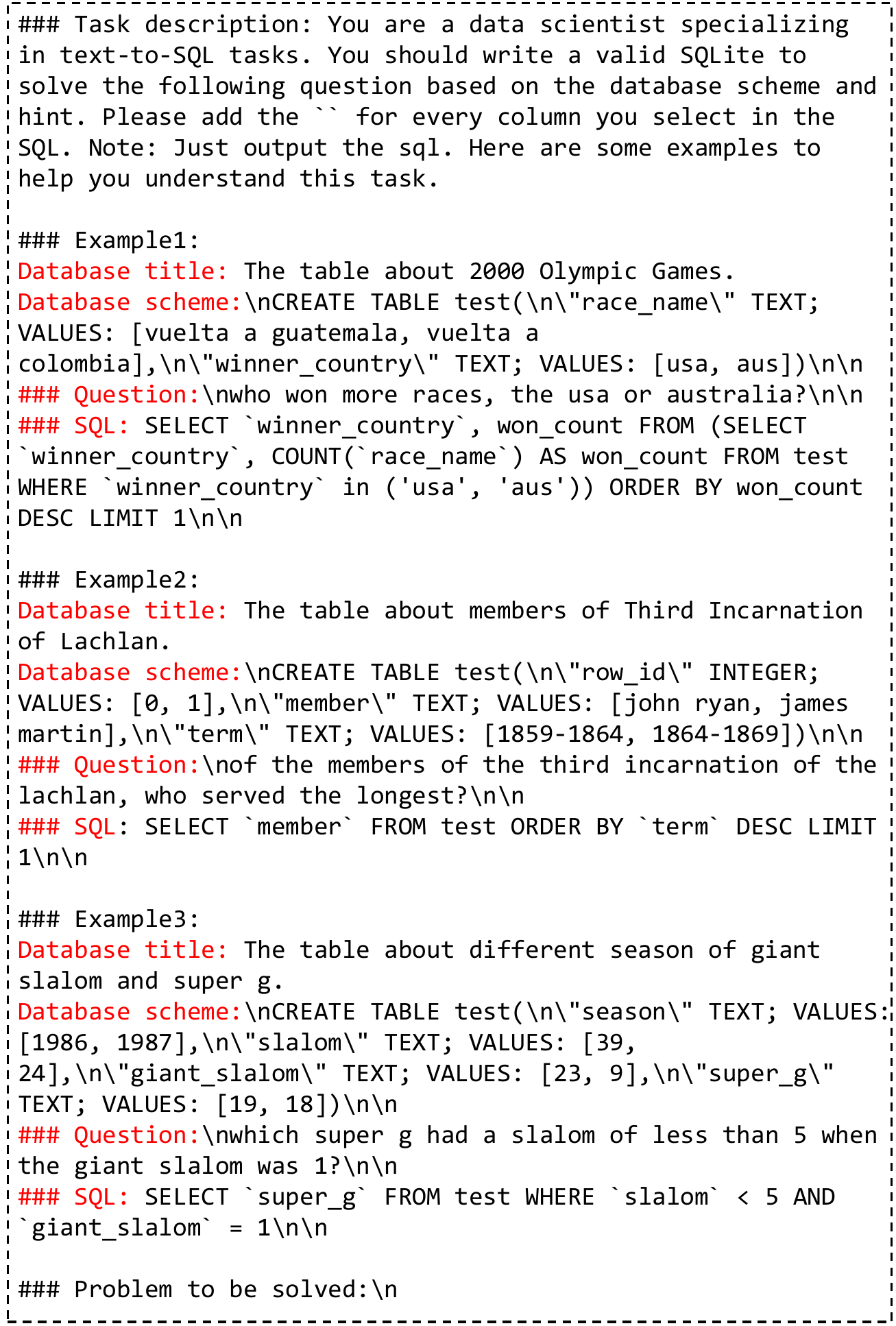}
\caption{Prompt for LLMs to generate SQL w/o CoT.}
\label{fig:sql_generation_wo_cot}
\end{figure}

\subsection{LLM Hyper-parameters}
\label{sec:appendix_hyper}
For all the procedures described in Section~\ref{sec:programtab}, we set the same hyper-parameters for LLMs.
Specifically, the temperature is set to 0.6 while both top\_p and the sample number are 1.

\subsection{Table Size Reduction}
\begin{figure}[t]
\centering
\includegraphics[scale=0.46]{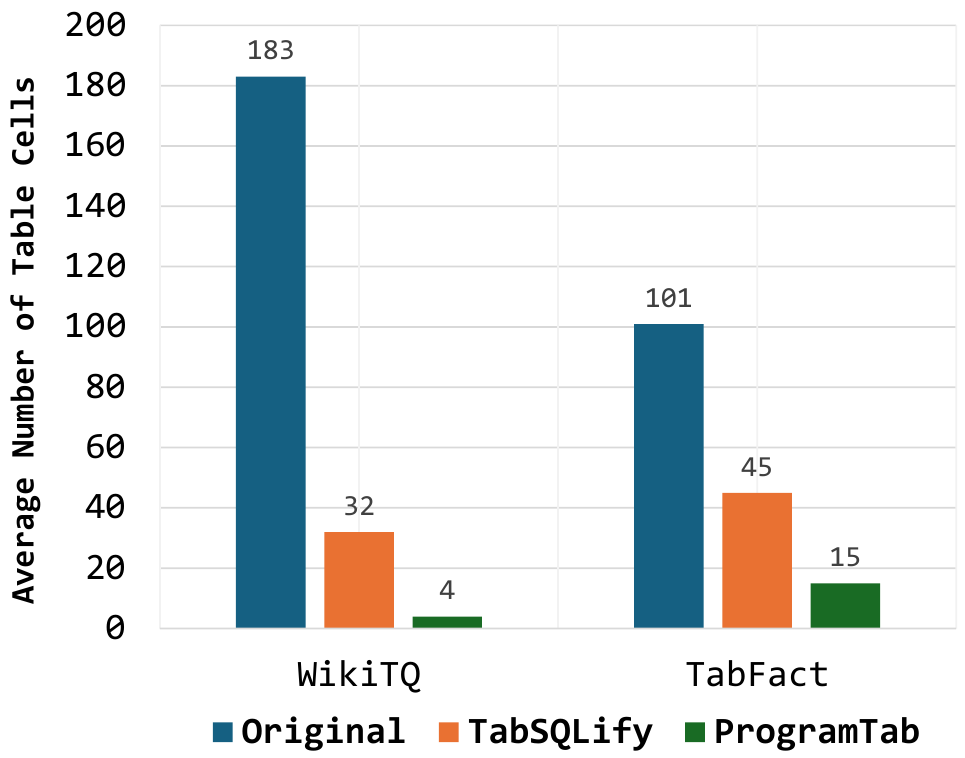}
\caption{Comparison of the average number of
table cells on two datasets.}
\label{fig:side_reduce}
\end{figure}

We analyze the efficiency of ProgramTab in filtering irrelevant information and extracting the key tabular data from tables.
To accomplish this, we count the average number of table cells that feed LLMs to generate the final answers.
As presented in Figure~\ref{fig:side_reduce}, the average number of full table cells (original) is 183 and 101 respectively.
There is a significant reduction after employing the TabSQLify approach.
Our framework ProgramTab employs SQL generation procedure to effectively filter much irrelevant information and extract the most related data. 
It respectively reduces the average number of table cells to 4 and 15 for WikiTQ and TabFact datasets.
These results also verify that ProgramTab can perform critical information extraction from amounts of tabular cells, and validly deal with large tables.

\subsection{Performance of Row Retrieval}
\label{sec:appendix_row}

We evaluate the performance of row retrieval stage in ProgramTab. 
Specifically, the average number of rows per table is 26 in the WikiTQ dataset. In the above experiments, we set K to 10, which means that on average, we can filter out 16 rows per table, resulting in a filtering ratio of 61.5\%. 
Additionally, among the top 10 rows retrieved, 86\% of the samples contained the final answer.

\subsection{Experiments on HiTab}
\label{sec:appendix_hitab}

\begin{table}[t]
\centering
\small
\renewcommand\arraystretch{1.0}
\resizebox{0.48\textwidth}{!}{
\begin{tabular}{lcc}
\toprule
\textbf{Methods} & \textbf{Backbone} & \textbf{HiTab}\\
\midrule
ReAct~\cite{yao2023react} & GPT-4 & 81.87 \\
$E^5$~\cite{zhang-etal-2024-e5} &  GPT-4 & 85.08 \\
\textbf{ProgramTab} &  GPT-4o-mini & \textbf{83.57} \\
\bottomrule
\end{tabular}
}
\caption{Performance of ProgramTab on HiTab dataset.}
\label{tb.hitab}
\end{table}

To present the effectiveness of ProgramTab when applied to more complex tabular structures,
we supplemented ProgramTab's experiments on the HiTab dataset, which contains hierarchical tables. To achieve this, we first reconstructed the hierarchical tables by merging certain column header information using a ":" delimiter, making them more suitable for processing by ProgramTab. The final experimental results are as follows: ProgramTab with GPT-4o-mini demonstrates promising performance, while adopting the $E^5$ method to process hierarchical tables yields even better performance.

\end{document}